\documentclass{article}

\usepackage{arxiv}

\usepackage[utf8]{inputenc} % allow utf-8 input
\usepackage[T1]{fontenc}    % use 8-bit T1 fonts
\usepackage{hyperref}       % hyperlinks
\usepackage{url}            % simple URL typesetting
\usepackage{booktabs}       % professional-quality tables
\usepackage{amsfonts}       % blackboard math symbols
\usepackage{nicefrac}       % compact symbols for 1/2, etc.
\usepackage{microtype}      % microtypography
\usepackage{lipsum}
\usepackage{array,multirow}

\usepackage{caption}
\usepackage{threeparttable}
\usepackage [english]{babel}
\usepackage [autostyle, english = american]{csquotes}
\MakeOuterQuote{"}

\title{Using Cognitive Models to Train Warm Start Reinforcement Learning Agents for Human-Computer Interactions\thanks{This is a preprint of our position paper presented to the "Reinforcement Learning for Humans, Computer, and Interaction (RL4HCI)" workship at ACM CHI2021, \url{https://sites.google.com/view/rl4hci/home}. The preprint is published under the Creative Commons  Attribution 4.0 International (CC BY 4.0)  license.}}

\author{
  Chao Zhang\thanks{These authors contributed equally to this research.} \\
  Department of Psychology\\
  Utrecht University\\
  Utrecht, The Netherlands \\
  \texttt{c.zhang@uu.nl} \\
   \And
 Shihan Wang\footnotemark[2] \\
  Department of Information and Computing Sciences\\
  Utrecht University\\
  Utrecht, The Netherlands \\
  \texttt{s.wang2@uu.nl } \\
  \And
 Henk Aarts \\
  Department of Psychology\\
  Utrecht University\\
  Utrecht, The Netherlands \\
  \texttt{h.aarts@uu.nl } \\
  \And
 Mehdi Dastani \\
  Department of Information and Computing Sciences\\
   Utrecht University\\
  Utrecht, The Netherlands \\
  \texttt{m.m.dastani@uu.nl } \\
}
\usepackage{graphicx}
\graphicspath{ {./} }

\begin{document}
\maketitle

\begin{abstract}
Reinforcement learning (RL) agents in human-computer interactions applications require repeated user interactions before they can perform well. To address this "cold start" problem, we propose a novel approach of using cognitive models to pre-train RL agents before they are applied to real users. After briefly reviewing relevant cognitive models, we present our general methodological approach, followed by two case studies from our previous and ongoing projects. We hope this position paper stimulates conversations between RL, HCI, and cognitive science researchers in order to explore the full potential of the approach. 
\end{abstract}

% keywords can be removed

\section{Introduction}
%Popularity of applying RL in HCI

%Strengths of RL and contrasts with other AI approaches

%Limitations of RL and the "cold start" problem in particular

%Objective of the position paper: Introducing a novel approach of using cognitive models to pre-train RL agents

%Structure of the position paper
%\section{The "Cold Start" of RL Agents}
%A more detailed description of the problem (why it undermines meaningful user interactions) and briefly review existing approaches to tackle the problem

%To what extend similar approaches has been tried before

%Reinforcement learning (see e.g. [7,9]) is a natural choice for personalization for health and wellbeing as it focuses on selecting the best actions with a more long term reward in mind. Here, the strategy defining which action to perform in what user state is referred to as a policy.
Reinforcement learning (RL) has gained growing popularity in many human-computer interaction (HCI) applications \cite{scheffler2002automatic,bassen2020reinforcement,yom2017encouraging}. %,
In digital health interventions, for example, RL is a natural choice for personalization as RL agents can continuous adapt their strategies based on users' responses to the interventions \cite{yom2017encouraging}. Moreover, the recent advances in interactive RL calls for contributions from HCI researchers to improve the efficiency of RL algorithms \cite{arzate2020survey}. 

While there is a natural fit between RL and HCI, the well-known data greedy property of reinforcement learning makes the RL-based systems often suffer from the cold start problem \cite{sutton2018reinforcement}. In HCI, as very few (or even no) experiences with users are available at the beginning in general, RL agents are required to interact many times with users prior to performing well. Many researchers had made efforts to overcome this challenge by shortening the learning process. Several approaches have been proposed to perform a faster online learning so that less interactions are needed in practice. For instance, Tabatabaei et al. \cite{tabatabaei2018narrowing} and Tomkins et al. \cite{tomkins2019intelligent} make RL algorithms quickly learn from the limited experience at the beginning stage by considering similar users. Gonul et al. \cite{gonul2018optimization} transfer the common knowledge acquired in other environments to get a faster convergence. Alternatively, some researches introduced prior knowledge from historical data and learn an initial policy offline in advance \cite{liao2020personalized,ameko2020offline}. It is also known as the \textit{warm start} RL agents \cite{silva2019neural}. 

The significance of such \textit{cold start} problem can be expanded for developing meaningful and useful interactive systems. First, while user experience is highly appreciated in HCI systems, online faster learning approaches might still bother users by too many interactions during collecting enough experience for learning. Second, the 'warm start' RL agents require the feedback from users for very specific tasks, which makes the collection of historical data very expensive. In many cases, the historical data may miss counterfactual information (i.e. what would have been the outcome had interventions or circumstances been different). 

To address the above issues, we propose to use cognitive models to complement the missing information for training \textit{warm start} RL agents for HCI applications. Recent advances in psychology and cognitive sciences have offered researchers many modeling tools that can be used for simulate the behaviors of human users in complex task environments. While training RL agents using simulation data is not new \cite{scheffler2002automatic,todi2021adapting}, we believe that RL researchers are yet to harnesses the potentials of adopting mature modeling tools from cognitive psychology. Before proposing a general methodology of using these models to train RL agents, we briefly review some of the most popular models.

\section{Simulate User Behavior Using Cognitive Models}
\label{cognitive}
\subsection{Human Reinforcement Learning}
RL was historically inspired by psychology and neuroscience, so it is no wonder that many modern RL algorithms are also used for modeling human learning \cite{sutton2018reinforcement}. A prominent example is the discovery that human brain uses a mechanism similar to temporal difference learning to update reward expectations, as in the classical Pavlovian conditioning \cite{o2003temporal}. In terms of more complex human behaviors, goal-directed learning and habit learning have also been successfully modeled using model-based and model-free reinforcement learning respectively \cite{daw2005uncertainty}. For HCI applications, RL models are especially useful to model how users learn to make better decisions in recurrent choice environments, where rewards are often temporarily discounted \cite{fu2006recurrent}. In general, modeling both user behavior and agent's behavior using RL creates an intriguing multi-agent RL problem, which is yet to be fully explored (e.g., \cite{peng2020understanding}).   

\subsection{Evidence Accumulation Models}
Evidence accumulation models (EAM), also known as sequential sampling models, are a class of models that explain human decision-making as a process of sampling and accumulating evidence before committing to a choice. Original developed for modeling memory retrieval \cite{ratcliff1978theory} and perceptual decisions \cite{purcell2010neurally}, EAMs have recently been adpated to model value-based decision-making tasks \cite{busemeyer2019cognitive}, such as food choices \cite{milosavljevic2010drift} and consumer purchase decisions \cite{krajbich2012attentional}. A unique value of EAMs is that by modeling users' choice and decision time data, their cognitive states can be estimated (e.g., decision threshold, preferences, and biases), which are otherwise unobservable \cite{milosavljevic2010drift}. These hidden states can be of interests in HCI applications, for example, as targets for intervention in behavior change support systems \cite{zhang2019towards}. These states may also be potentially incorporated into reward functions for RL agents if the goal is to change cognitive states rather than overt behaviors. 

\subsection{Cognitive Architectures}
Unlike RL and EAMs, which model specific aspects of human cognition (i.e., learning and decision-making), cognitive architectures are general computational frameworks that simulate how human brain produces thoughts, language, and actions. Notable examples of cognitive architectures include the Adaptive Control of Thought - Rational \cite{anderson2004integrated}, Soar \cite{laird2012soar}, and the BOID architecture \cite{broersen2001boid}. With their abilities to emulate human perception, attention, and memory mechanisms, they are especially useful for modeling human-machine interactions in multi-task environments. For example, ACT-R and its variants have been used to simulate human driving behaviors, such as lane-keeping \cite{salvucci2001toward} and car-following \cite{deng2019modeling}. ACT-R has also been used to model menu search in HCI \cite{bailly2014model,hornof1997cognitive} and recently simulated search data have been used to train RL agents \cite{todi2021adapting}.

%ACT-R was also used in our ongoing project described in Section 5.2. 

\section{Train RL Agents by User Simulator}
%Discuss a general methodological approach; what contributions our approach can make for related fields, RL research, cognitive modeling, and applying RL in HCI. 

\begin{figure*}[ht]
  \centering
  \includegraphics[width=.9\linewidth]{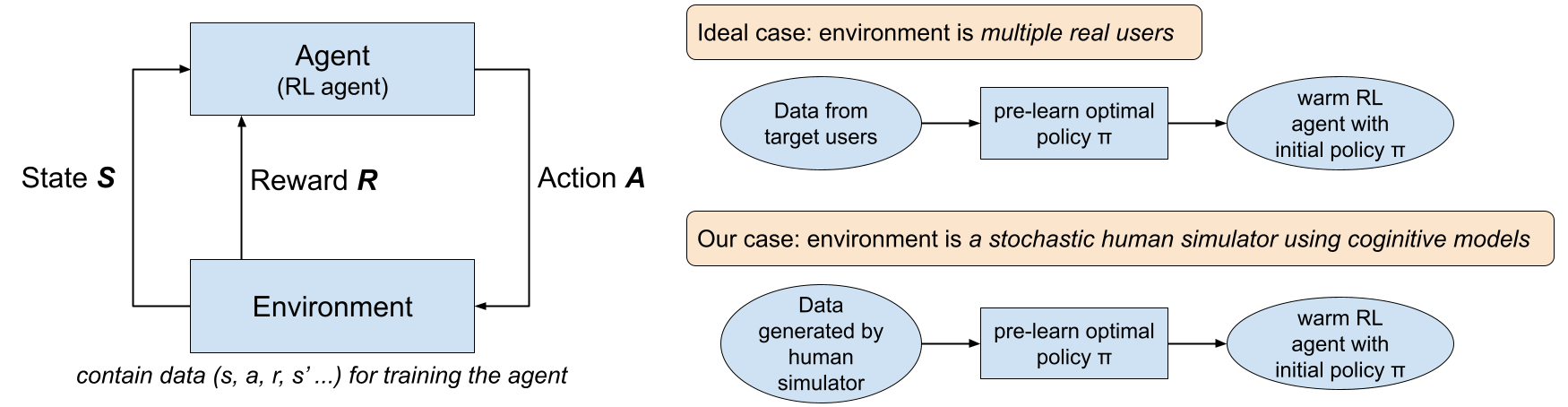}
  \caption{The overview of our methodology, including the agent–environment interaction in a given MDP model and two cases for pre-learning the initial policy for warm start RL agents. }
  \label{fig:rl}
\end{figure*}

Many practical HCI problems can be formalized as a finite horizon and discrete time Markov Decision Making (MDP) tasks. We present an overview of agent-environment (computer-human) interaction in Figure \ref{fig:rl}. Here
the agent represents any intelligent and interactive HCI system that interacts with a target user (which is the environment). At each time step, the RL agent observes a contextual representation of the environment, and on that basis selects an action. Afterwards, the environment passes a numerical reward (inferring users' feedback on the given action) back to the agent and updates itself in a new state. %In this article, we aim to obtain a warm start RL agent by learning an initial policy offline from prior knowledge and historical data. 
To achieve this goal, ideally, we could collect a set of historical data from real target users, then feed the data into the interaction loop between agent and environment. Based on the trial-and-error mechanism, the RL agent could estimate its optimal policy for maximizing our expected long-term reward. However, as discussed above, such data collection for specific HCI tasks can be very expensive \cite{liao2020personalized}. Our method therefore aims to overcome the situations that is not possible to collect the interactive data directly from the users (i.e. due to restrictions of interactions like when a wrong interaction may cause serious problems). In particular, we follow the second case shown in Figure \ref{fig:rl} and develop a stochastic human simulator based on cognitive models for generating the interactive user data. As mentioned in section \ref{cognitive}, several cognitive models could be utilized to model the human learning and decision making procedures, which could imply the upcoming behaviors (next states) and feedback (rewards) of target users. In this manner, we can generate the required data from the human simulator and pre-learn an optimal initial policy for the warm start RL agent. 

%\todo{Not sure how to answer:what contributions our approach can make for related fields? And I do think this shall not be filled in this section}

\section{Case Studies}
\subsection{Promoting Physical Activity}
In one of our earlier projects, the goal of the RL agent was to use mobile notifications to motivate physical exercises, with the constrain that only a limited number of notifications could be sent to the users \cite{wang2021optimizing}. A large data set of running behaviors was accessible, but it lacked the information about the contexts when users decided not to run and the effects of notifications. To pre-train our agent, we simulated users' running decisions using a Dynamic Bayesian network that combined the historical running data and psychological theories about memory and decision-making. Our results indicated that the RL agent pre-trained using this approach could deliver notifications in a manner that realizes higher behavioral impact than context-blind agents. We are currently conducting user studies to further validate our method. 

\subsection{Intelligent Driving Assist}
%Summarize our ongoing project
In an ongoing project, we are developing a driving assist that helps drivers to keep lanes. A desirable characteristic is that the intelligent assist should only intervene in situations where the human driver is distracted from the driving task, in order to preserve the autonomy of the driver. To achieve this goal, the first step is to use ACT-R to simulate a human driver in a simple high-way scenario, where discrete distracting events would occur. With the simulate driving data, 
we pre-train an RL agent and find the best RL algorithm in simulation experiments. Finally, we plan to conduct an human experiment with a driving simulator to evaluate both the cognitive model and the effectiveness of the RL-based driving assist. 

\section{Concluding Remarks}
%limitations; Outlook for future research; stimulating more interests in this approach
We propose a methodology to use cognitive models to train \textit{warm start} RL agents for HCI applications. While our previous works have shown some promises of our approach, two caveats are worth noting when applying our approach to HCI applications. First, all the cognitive models we introduced by default can simulate how an average human being behaves, but accounting for individual differences is more challenging. In HCI applications where user personalization is crucial, our approach should be strengthened either by matching parameter values used in the model simulation with the measured parameter values from the real user, or simply by continuing RL based on repeated interactions with the real users. Nonetheless, an RL agent that is able to deal with an average user is already a lot better than a "cold-start" agent.

Second, pre-training RL agents using cognitive models creates a dependency. In a word, the effectiveness of an RL agent is bounded to the validity of the cognitive model used for its training. Moreover, in cases where the RL agent also influence users' cognition and behaviors, additional assumptions may be required to be built into the cognitive model. Therefore, we recommend to use empirically validated cognitive models whenever possible, and otherwise untested assumptions must be stated. On the positive side, our approach does provide an interesting way to test cognitive models using user experiments where trained RL agents are also at play. 

\section*{Acknowledgement}
Two case studies in this paper are supported by NWO \& SIA (Grant Number 629.004.013) and the seed fund of SIGs “Autonomous Intelligent Systems” and “Social and Cognitive Modeling” at \href{https://www.uu.nl/en/research/human-centered-artificial-intelligence}{the focus area \textit{Human-centered AI} of Utrecht University} respectively. The contributions from Chao Zhang and Henk Aarts are supported by \href{https://human-ai.nl/}{\textit{the Alliance project HUMAN-AI}} funded by Utrecht University, Eindhoven University of Technology, Wageningen University \& Research, and University Medical Center Utrecht.

\bibliographystyle{unsrt} 
\bibliography{RL4HCI} 

\begin{thebibliography}{10}

\bibitem{scheffler2002automatic}
Konrad Scheffler and Steve Young.
\newblock Automatic learning of dialogue strategy using dialogue simulation and
  reinforcement learning.
\newblock In {\em Proceedings of HLT}, volume~2. Citeseer, 2002.

\bibitem{bassen2020reinforcement}
Jonathan Bassen, Bharathan Balaji, Michael Schaarschmidt, Candace Thille, Jay
  Painter, Dawn Zimmaro, Alex Games, Ethan Fast, and John~C Mitchell.
\newblock Reinforcement learning for the adaptive scheduling of educational
  activities.
\newblock In {\em Proceedings of the 2020 CHI Conference on Human Factors in
  Computing Systems}, pages 1--12, 2020.

\bibitem{yom2017encouraging}
Elad Yom-Tov, Guy Feraru, Mark Kozdoba, Shie Mannor, Moshe Tennenholtz, and
  Irit Hochberg.
\newblock Encouraging physical activity in patients with diabetes: intervention
  using a reinforcement learning system.
\newblock {\em Journal of medical Internet research}, 19(10):e338, 2017.

\bibitem{arzate2020survey}
Christian Arzate~Cruz and Takeo Igarashi.
\newblock A survey on interactive reinforcement learning: Design principles and
  open challenges.
\newblock In {\em Proceedings of the 2020 ACM Designing Interactive Systems
  Conference}, pages 1195--1209, 2020.

\bibitem{sutton2018reinforcement}
Richard~S Sutton and Andrew~G Barto.
\newblock {\em Reinforcement learning: An introduction}.
\newblock MIT press, 2018.

\bibitem{tabatabaei2018narrowing}
Seyed~Amin Tabatabaei, Mark Hoogendoorn, and Aart van Halteren.
\newblock Narrowing reinforcement learning: Overcoming the cold start problem
  for personalized health interventions.
\newblock In {\em International Conference on Principles and Practice of
  Multi-Agent Systems}, pages 312--327. Springer, 2018.

\bibitem{tomkins2019intelligent}
Sabina Tomkins, Peng Liao, Serena Yeung, Predrag Klasnja, and Susan Murphy.
\newblock Intelligent pooling in thompson sampling for rapid personalization in
  mobile health.
\newblock 2019.

\bibitem{gonul2018optimization}
Suat Gonul, Tuncay Namli, Mert Baskaya, Ali~Anil Sinaci, Ahmet Cosar, and
  Ismail~Hakki Toroslu.
\newblock Optimization of just-in-time adaptive interventions using
  reinforcement learning.
\newblock In {\em International Conference on Industrial, Engineering and Other
  Applications of Applied Intelligent Systems}, pages 334--341. Springer, 2018.

\bibitem{liao2020personalized}
Peng Liao, Kristjan Greenewald, Predrag Klasnja, and Susan Murphy.
\newblock Personalized heartsteps: A reinforcement learning algorithm for
  optimizing physical activity.
\newblock {\em Proceedings of the ACM on Interactive, Mobile, Wearable and
  Ubiquitous Technologies}, 4(1):1--22, 2020.

\bibitem{ameko2020offline}
Mawulolo~K Ameko, Miranda~L Beltzer, Lihua Cai, Mehdi Boukhechba, Bethany~A
  Teachman, and Laura~E Barnes.
\newblock Offline contextual multi-armed bandits for mobile health
  interventions: A case study on emotion regulation.
\newblock In {\em Fourteenth ACM Conference on Recommender Systems}, pages
  249--258, 2020.

\bibitem{silva2019neural}
Andrew Silva and Matthew Gombolay.
\newblock Neural-encoding human experts' domain knowledge to warm start
  reinforcement learning.
\newblock {\em arXiv preprint arXiv:1902.06007}, 2019.

\bibitem{todi2021adapting}
Kashyap Todi, Luis~A Leiva, Gilles Bailly, and Antti Oulasvirta.
\newblock Adaptive user interfaces with model-based reinforcement learning.
\newblock In {\em Proceedings of the 2021 CHI Conference on Human Factors in
  Computing Systems}, 2021.

\bibitem{o2003temporal}
John~P O'Doherty, Peter Dayan, Karl Friston, Hugo Critchley, and Raymond~J
  Dolan.
\newblock Temporal difference models and reward-related learning in the human
  brain.
\newblock {\em Neuron}, 38(2):329--337, 2003.

\bibitem{daw2005uncertainty}
Nathaniel~D Daw, Yael Niv, and Peter Dayan.
\newblock Uncertainty-based competition between prefrontal and dorsolateral
  striatal systems for behavioral control.
\newblock {\em Nature neuroscience}, 8(12):1704--1711, 2005.

\bibitem{fu2006recurrent}
Wai-Tat Fu and John~R Anderson.
\newblock From recurrent choice to skill learning: A reinforcement-learning
  model.
\newblock {\em Journal of experimental psychology: General}, 135(2):184, 2006.

\bibitem{peng2020understanding}
Zhinan Peng, Jiangping Hu, Yiyi Zhao, and Bijoy~K Ghosh.
\newblock Understanding the mechanism of human--computer game: a distributed
  reinforcement learning perspective.
\newblock {\em International Journal of Systems Science}, 51(15):2837--2848,
  2020.

\bibitem{ratcliff1978theory}
Roger Ratcliff.
\newblock A theory of memory retrieval.
\newblock {\em Psychological review}, 85(2):59, 1978.

\bibitem{purcell2010neurally}
Braden~A Purcell, Richard~P Heitz, Jeremiah~Y Cohen, Jeffrey~D Schall, Gordon~D
  Logan, and Thomas~J Palmeri.
\newblock Neurally constrained modeling of perceptual decision making.
\newblock {\em Psychological review}, 117(4):1113, 2010.

\bibitem{busemeyer2019cognitive}
Jerome~R Busemeyer, Sebastian Gluth, J{\"o}rg Rieskamp, and Brandon~M Turner.
\newblock Cognitive and neural bases of multi-attribute, multi-alternative,
  value-based decisions.
\newblock {\em Trends in cognitive sciences}, 23(3):251--263, 2019.

\bibitem{milosavljevic2010drift}
Milica Milosavljevic, Jonathan Malmaud, Alexander Huth, Christof Koch, and
  Antonio Rangel.
\newblock The drift diffusion model can account for value-based choice response
  times under high and low time pressure.
\newblock {\em Judgment and Decision Making}, 5(6):437--449, 2010.

\bibitem{krajbich2012attentional}
Ian Krajbich, Dingchao Lu, Colin Camerer, and Antonio Rangel.
\newblock The attentional drift-diffusion model extends to simple purchasing
  decisions.
\newblock {\em Frontiers in psychology}, 3:193, 2012.

\bibitem{zhang2019towards}
Chao Zhang.
\newblock {\em Towards a psychological computing approach to digital lifestyle
  interventions}.
\newblock PhD Dissertation. Eindhoven University of Technology, 2019.

\bibitem{anderson2004integrated}
John~R Anderson, Daniel Bothell, Michael~D Byrne, Scott Douglass, Christian
  Lebiere, and Yulin Qin.
\newblock An integrated theory of the mind.
\newblock {\em Psychological review}, 111(4):1036, 2004.

\bibitem{laird2012soar}
John~E Laird.
\newblock {\em The Soar cognitive architecture}.
\newblock MIT press, 2012.

\bibitem{broersen2001boid}
Jan Broersen, Mehdi Dastani, Joris Hulstijn, Zisheng Huang, and Leendert
  van~der Torre.
\newblock The boid architecture: conflicts between beliefs, obligations,
  intentions and desires.
\newblock In {\em Proceedings of the fifth international conference on
  Autonomous agents}, pages 9--16, 2001.

\bibitem{salvucci2001toward}
Dario~D Salvucci, Erwin~R Boer, and Andrew Liu.
\newblock Toward an integrated model of driver behavior in cognitive
  architecture.
\newblock {\em Transportation Research Record}, 1779(1):9--16, 2001.

\bibitem{deng2019modeling}
Chao Deng, Chaozhong Wu, Shi Cao, and Nengchao Lyu.
\newblock Modeling the effect of limited sight distance through fog on
  car-following performance using qn-actr cognitive architecture.
\newblock {\em Transportation research part F: traffic psychology and
  behaviour}, 65:643--654, 2019.

\bibitem{bailly2014model}
Gilles Bailly, Antti Oulasvirta, Duncan~P Brumby, and Andrew Howes.
\newblock Model of visual search and selection time in linear menus.
\newblock In {\em Proceedings of the sigchi conference on human factors in
  computing systems}, pages 3865--3874, 2014.

\bibitem{hornof1997cognitive}
Anthony~J Hornof and David~E Kieras.
\newblock Cognitive modeling reveals menu search in both random and systematic.
\newblock In {\em Proceedings of the ACM SIGCHI Conference on Human factors in
  computing systems}, pages 107--114, 1997.

\bibitem{wang2021optimizing}
Shihan Wang, Chao Zhang, Ben Kröse, and Herke van Hoof.
\newblock Optimizing adaptive notifications in mobile healthinterventions
  systems: Reinforcement learning froma data-driven behavioral simulator.
\newblock Manuscript Under Review.

\end{thebibliography}
\end{document}